\documentclass[twoside,11pt]{article} % 
\usepackage{jmlr2e}
\usepackage{times}
\usepackage{amsmath}
\usepackage{multirow}
\usepackage{booktabs}
\usepackage{hyperref,url,dsfont,nicefrac}
\hypersetup{
    colorlinks,
    breaklinks,
    linkcolor = blue,
    citecolor = blue,
    urlcolor  = blue,
}
\newcommand{\nop}[1]{}
\usepackage{wrapfig}
\usepackage{mathrsfs}
\usepackage{epsfig}
\usepackage{graphics}
\usepackage{subfigure}
\usepackage{epsf}
\usepackage{soul}
\usepackage{nicefrac}       % compact symbols for 1/2, etc.
\usepackage{microtype}      % microtypography

\usepackage{algorithm,algorithmic}
\newtheorem{ass}{Assumption}
\newtheorem{thm}{Theorem}

\newtheorem{lem}{Lemma}
\usepackage{dsfont}

\def \D {\mathcal{D}}

\def \w {\mathbf{w}}

\def \R {\mathbb{R}}

\def \W {\mathcal{W}}

\def \A {\mathcal{A}}

\def \B {\mathbf{B}}
\def \tb {\tilde{b}}

\def \tg {\tilde{g}}

\def \MS {\mathcal{S}}

\def \sumT {\sum_{t=1}^T}

\def \v {\mathbf{v}}

\def \B {\mathcal{B}}

\def \MS {\mathcal{S}}

\def \ind {\mathbb{I}}

\DeclareMathOperator*{\Reg}{Regret}

\DeclareMathOperator*{\DReg}{D-Regret}

\DeclareMathOperator*{\sgn}{sign}
\DeclareMathOperator*{\erf}{erf}

\usepackage{graphicx,color} % more modern

\usepackage{mathtools}

\begin{document}

\title{Discounted Online Convex Optimization: Uniform  Regret\\  Across a Continuous Interval}

\author{\name Wenhao Yang \email yangwh@lamda.nju.edu.cn \\
       \name Sifan Yang \email yangsf@lamda.nju.edu.cn \\
       \name Lijun Zhang \email zhanglj@lamda.nju.edu.cn \\
       \addr National Key Laboratory for Novel Software Technology, Nanjing University, China\\
       \addr School of Artificial Intelligence, Nanjing University, China
       }

\maketitle

\begin{abstract}%
Reflecting the greater significance of recent history over the distant past in non-stationary environments, $\lambda$-discounted regret has been introduced in online convex optimization (OCO) to gracefully forget past data as new information arrives. When the discount factor $\lambda$ is given, online gradient descent with an appropriate step size achieves an $O(1/\sqrt{1-\lambda})$ discounted regret. However, the value of $\lambda$ is often not predetermined in real-world scenarios. This gives rise to a significant \emph{open question}: is it possible to develop a discounted algorithm that adapts to an unknown discount factor. In this paper, we affirmatively answer this question by providing a novel analysis to demonstrate that smoothed OGD (SOGD)  achieves a uniform $O(\sqrt{\log T/1-\lambda})$ discounted regret, holding for all values of $\lambda$ across a continuous interval simultaneously. The basic idea is to maintain multiple OGD instances to handle different discount factors, and aggregate their outputs sequentially by an online prediction algorithm named as Discounted-Normal-Predictor (DNP)~\citep{DNP:2010:Arxiv}. Our analysis reveals that DNP can combine the decisions of two experts, even when they operate on discounted regret with different discount factors.
\end{abstract}

% \begin{keywords}%
%   Online Convex Optimization, Discounted Regret, Adaptive Algorithms%
% \end{keywords}

\section{Introduction}
Online convex optimization (OCO) serves as a fundamental framework for online learning, effectively modeling many real-world  problems~\citep{Intro:Online:Convex}. OCO can be viewed as a repeated game between the learner and the environment, governed by the following protocol. In each round $t\in [T]$, the learner chooses a decision $\w_t$ from a convex domain $\W\subseteq \R^d$. Then, the learner suffers a loss $f_t(\w_t)$ and observe some information, where $f_t\colon \W \mapsto \R $ is chosen by the environment. To evaluate the performance of the learner, static regret is commonly used~\citep{bianchi-2006-prediction}:
\begin{equation*}
    \Reg (T) = \sumT f_t(\w_t) - \min_{\w\in\W}\sumT f_t(\w)
\end{equation*}
which is defined as the difference between the cumulative loss of the online learner and that of the best decision chosen in hindsight. However, static regret is not well-suited for changing environments where the future significantly diverges from the past. To facilitate gracefully forgetting past data as new information arrives, $\lambda$-discounted regret has been proposed~\citep{arxiv:2008:Freund}: 
\begin{equation} \label{eqn:dis:regret}
 \DReg (T,\lambda)= \sum_{t=1}^T  \lambda^{T-t}  f_t(\w_t)  -  \min_{\w \in \W} \sum_{t=1}^T  \lambda^{T-t}  f_t(\w) 
\end{equation}
where $\lambda \in (0,1)$ is the discount factor, denoting the degree of forgetting of the past. 
% A smaller value of $\lambda$ indicates a greater emphasis on forgetting past data. 

Although discounted regret has been explored to some extent in prediction with expert advice (PEA)~\citep{arxiv:2008:Freund,ALT:2010:Chernov,DNP:2010:Arxiv,NIPS:2012:Cesa-bianchi,ICML:2014:Krichene} and games~\citep{AAAI:2019:Brown,ICLR:2024:Xu}, discounted OCO has been relatively underexplored in the literature. Recently, \citet{pmlr-v235-zhang24e} studied a more challenging setting with time-varying discount factors, aiming to establish gradient-adaptive regret bounds. For the special case of a fixed factor, they establish an $O(1/\sqrt{1-\lambda^2})$ result for $\lambda$-discounted regret, but their algorithm requires the prior knowledge of discount factor $\lambda$. However, in non-stationary environments, the appropriate discount factor may change over time or vary depending on the specific task. Taking stock prediction as an example, the importance of past data varies with market conditions. During volatile periods, recent data becomes more relevant, requiring a smaller discount factor, while in stable periods, a larger discount factor is more appropriate. Recognizing the considerable significance of adapting to an unknown discount factor in practical scenarios, they explicitly leave this scenario as an important \emph{open question} in their work. 

Notably, there exists another performance measurement for dealing with changing environments, known as adaptive regret. Defined as the maximal static regret for every interval $[r,s]\subseteq [T]$ over the whole time horizon, it serves a similar purpose  to that of discounted regret, as both effectively define a temporal horizon of interest. Specifically, the length of an interval in adaptive regret is conceptually analogous to the effective window size controlled by a discount factor; for instance, a smaller discount factor emphasizes recent data by promoting more rapid forgetting of the past, which corresponds to a focus on more recent intervals. Since existing algorithms for adaptive regret~\citep{Adaptive:Hazan,Improved:Strongly:Adaptive,Dynamic:Regret:Adaptive} provide guarantees that hold simultaneously for all intervals, it is natural to ask \emph{whether it is possible to design a discounted OCO algorithm that adapts to an unknown discount factor}. In this paper, we provide an affirmative answer.

\begin{figure}[t]
\begin{center}
\centerline{\includegraphics[width=.9\columnwidth]{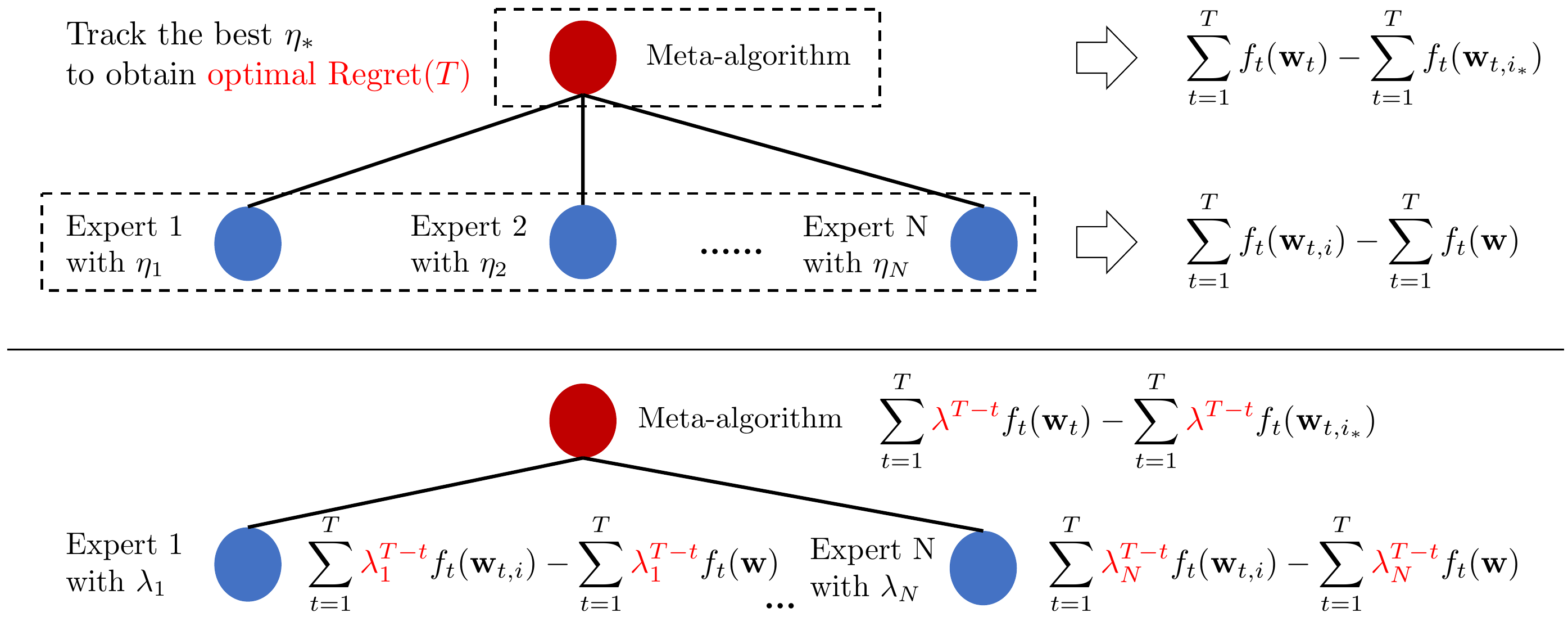}}
\caption{A meta-expert framework for OCO that adapts to unknown parameters (previous work, upper panel) and discounted OCO (our setting, lower panel). }
\label{fig:tc}
\end{center}
\end{figure}

\subsection{Technical Challenge} \label{sec:tc}
In the literature, extensive research has explored online algorithms that adapt to unknown parameters, including universal OCO~\citep{NIPS2016_6268,ICML:2022:Zhang}, dynamic regret~\citep{Adaptive:Dynamic:Regret:NIPS,pmlr-v134-baby21a}, and adaptive regret~\citep{Adaptive:ICML:15,Improved:Strongly:Adaptive}. These methods adopt the meta-expert framework as shown in the upper panel of Figure~\ref{fig:tc}, where they maintain multiple experts with different configurations and deploy a meta-algorithm to track the best one. Therefore, to adapt to an unknown discount factor, a straightforward idea is to apply this meta-expert framework by constructing multiple OGD instances, each operating with a different potential discount factor, and then using a meta-algorithm to combine their decisions. However, this strategy fails to handle the discounted scenario, because existing meta-algorithms used in the these  studies, such as Hedge~\citep{FREUND1997119} or Fixed-Share~\citep{Herbster1998}, typically require that all experts and the meta-algorithm  operate under \emph{a unified performance measurement}. For discounted OCO, the fact that the uncertainty of the discount factor is tied to the performance measurement makes this requirement difficult to satisfy. As depicted in the lower panel of Figure~\ref{fig:tc}, experts configured for different discount factors are essentially operating under \emph{different performance measures}, i.e., $\lambda$-discounted regret with varying $\lambda$. This renders the traditional meta-expert framework incapable of effectively handling an unknown discount factor.

\begin{figure}[t]
\begin{center}
\centerline{\includegraphics[width=.7\columnwidth]{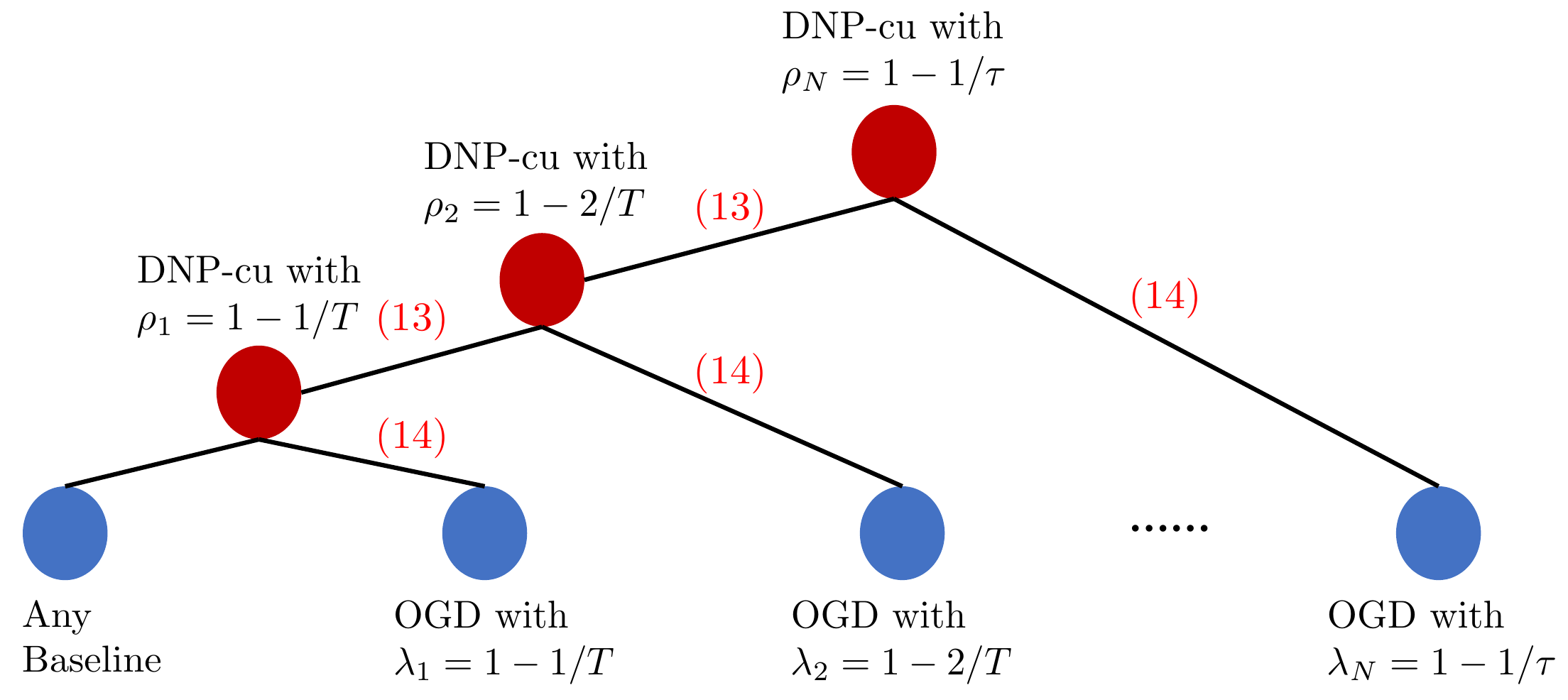}}
\caption{Overall procedure: sequentially aggregation by DNP-cu with different discount factors (red nodes) of OGD experts (blue nodes), using meta-regret  from  \eqref{eqn:alg2:lower3} and \eqref{eqn:alg2:lower4}.  }
\label{fig:discounted}
\end{center}
\end{figure}

\subsection{Our Solution and Contributions}
To address the above challenge, we revisit Smoothed OGD (SOGD)  \citep{Smooth:DNP:NeurIPS}, which is proposed to support the adaptive regret for smoothed OCO. The key idea of their work is to construct multiple instances of OGD with different step sizes, and employ Discounted-Normal-Predictor with conservative updating (DNP-cu)~\citep{DNP:2010:Arxiv,NIPS2011_11b921ef} as the meta-algorithm to sequentially aggregate their decisions. Following this idea, we first discretize the continuous interval of potential discount factors, say $\lambda\in [1-1/\tau,1-1/T]$ where $\tau$ is a minimal window length, by constructing a geometric series to cover the range of their values. For each possible discount factor $\lambda_i$, we create an expert by running an instance of OGD to achieve optimal $\lambda_i$-discounted regret. Then, we employ multiple instances of DNP-cu with different discount factors $\rho_i=\lambda_i$ to sequentially aggregate decisions from each expert. The overall procedure is illustrated in Figure~\ref{fig:discounted}. In this work, we analyze the performance of  DNP-cu under the discounted payoff setting, demonstrating its ability to effectively control the discounted regret. Furthermore, our novel analysis reveals that DNP-cu is able to successfully aggregate the decisions of two experts, \emph{even when they operate on discounted regret with different discount factors}. Significantly, our analysis establishes that this approach achieves a uniform $O(\sqrt{\log T/1-\lambda})$ bound for $\lambda$-discounted regret that holds  simultaneously for all $\lambda\in [1-1/\tau,1-1/T]$, and does not require knowing the value of $\lambda$. 

\paragraph{Clarification} Finally, we would like to emphasize that although the idea of deriving uniform discounted regret across a continuous interval has appeared in \citet{DNP:2010:Arxiv}, their analysis is conducted in the setting of PEA rather than OCO. Moreover, \citet[\S~5]{DNP:2010:Arxiv} only outlined the proof sketch without providing many of the technical details.  Reconstructing the complete argument and addressing the gaps and inaccuracies in their presentation requires significant effort. 

\paragraph{Organization} The rest of this work is organized as follows. 
Section~\ref{sec:pre} presents preliminaries and reviews mostly related works. Section~\ref{sec:main} presents main results for discounted OCO. Section~\ref{sec:ana} provides all the proofs and omitted details.  We finally conclude the
paper in Section~\ref{sec:con}. 

\section{Preliminaries and Related Works} \label{sec:pre}
In this section, we first present standard assumptions of OCO. Then, we review  related works to our paper, including discounted online learning and discounted-normal-predictor. 

\subsection{Preliminaries} 
We introduce the following common assumptions for OCO \citep{Online:suvery}.
\begin{ass}\label{ass:1} All the functions $f_t$'s are convex over the domain $\W$.
\end{ass}
\begin{ass}\label{ass:2} The gradients of all functions are bounded by $G$, i.e.,
\begin{equation}\label{eqn:grad}
\max_{\w \in \W}\|\nabla f_t(\w)\| \leq G, \ \forall t \in [T].
\end{equation}
\end{ass}
\begin{ass}\label{ass:3} The diameter of the domain $\W$ is bounded by $D$, i.e.,
\begin{equation}\label{eqn:domain}
\max_{\w, \w' \in \W} \|\w -\w'\| \leq D.
\end{equation}
\end{ass}
Without loss of generality, we assume \citep{Smooth:DNP:NeurIPS}
\begin{equation} \label{eqn:function:range}
f_t(\w) \in [0, GD], \ \forall \w \in \W, \ t \in [T].
\end{equation}

\subsection{Discounted Online Learning}
Due to the fact that recent information is often more important than past history in non-stationary environments, discounted online learning has been proposed to gradually forget the past as new data arrives. Under the setting of PEA, discounted regret is defined as~\citep{bianchi-2006-prediction}:
\begin{equation}\label{eqn:discount:another}
    \sumT \beta_{T-t} f_t(\w_t) - \min_{\w\in\W}\sumT \beta_{T-t} f_t(\w),
\end{equation}
where $\{\beta_t \}_{t=1}^T$ is a decreasing sequence of discount factors. The 
$\lambda$-discounted regret \eqref{eqn:dis:regret} studied in our work can be viewed as a special case of \eqref{eqn:discount:another}, also referred to as exponential discounting. The seminal work of~\citep{arxiv:2008:Freund} proposes a discounted variant of Hedge, which achieves an $O(\sqrt{\ln N/(1-\lambda)})$ regret bound where $N$ is the number of experts. Subsequent works~\citep{ALT:2010:Chernov,NIPS:2012:Cesa-bianchi,ICML:2014:Krichene} have also explored other discounted variants under the framework of ``tracking the best expert'', and proved bounds for discounted regret. Furthermore, \citet{AAAI:2019:Brown,ICLR:2024:Xu} propose discounted version of  counterfactual regret minimization (CFR) for solving imperfect-information games. Recently, discounted regret has gained attention in the context of OCO. \citet{pmlr-v235-zhang24e} investigate adaptive OGD and FTRL for discounted regret for discounted regret with time-varying factors. For online linear regression, \citet{ICML:2024:Jacobsen} present a discounted variant of the VAW forecaster, which enjoys dynamic regret guarantee. Furthermore, they also extend their results to strongly adaptive regret. Based on online-to-non-convex conversion~\citep{ICML:2023:Online:To:Nonconvex}, \citet{ICML:2024:Ahn} investigate Adam optimizer, and  propose a conversion from the discounted regret to the dynamic regret.  However, these aforementioned works cannot adapt to  an unknown discount factor. 

\begin{algorithm}[t]
\caption{Discounted-Normal-Predictor}
\begin{algorithmic}[1]
\REQUIRE Two parameters: $\rho$ and $Z$
\STATE Set $x_1=0$
\FOR{$t=1,\ldots,T$}
\STATE Predict $g(x_t)$
\STATE Receive $b_t$
\STATE Set $x_{t+1}=\rho x_t + b_t$
\ENDFOR
\end{algorithmic}
\label{alg:1}
\end{algorithm}

\subsection{Discounted-Normal-Predictor}
Discounted-Normal-Predictor (DNP)~\citep{DNP:2010:Arxiv} was introduced to solve the bit prediction problem, with its protocol described as follows. Consider an adversarial sequence of bits $b_1,\ldots,b_T$ where each $b_t \in [-1,1]$ can take real values, and our goal is to predict the next bit from the previous bits. In each round $t\in [T]$, the learner predicts a confidence $c_t \in [-1,1]$, and observes the value of $b_t$. Then, the learner gets a payoff $c_t b_t$ at each round. For a $T$-round game, the goal of the learner is to maximize the cumulative payoff $\sum_{t=1}^T c_t b_t$. 

The overall procedure of DNP is summarized in Algorithm~\ref{alg:1}. In each round $t$, DNP maintains a discounted deviation defined as $x_t = \sum_{j=1}^{t-1} \rho^{t-1-j} b_j$, where the discount factor $\rho=1-1/n$ and $n >0$ is a parameter for the interval length. In Step~3, DNP predicts the confidence level $g(x_t)$ by a confidence function $g(\cdot)$, which is defined as 
\begin{equation} \label{eqn:confidence:original}
g(x)=\sgn(x) \cdot \min \left( Z \cdot \erf\left( \frac{|x|}{4 \sqrt{n}} \right) e^{\frac{x^2}{16 n}}, 1\right)
\end{equation}
where $Z>0$ is a parameter, and $\erf(x) = \frac{2}{\sqrt{\pi}} \int_0^x e^{-s^2} d s$ is the error function. For any $Z \leq 1/e$, \citet[Theorem 14]{DNP:2010:Arxiv} have proved that DNP satisfies
\begin{equation}\label{eqn:DNP:main:thm}
    \sum_{t=1}^T g(x_t) b_t  \geq \max\left( \left|\sum_{j=1}^T b_j \right| -O \big(\sqrt{T \log (1/Z)}\big), - O\big(Z\sqrt{T}\big)\right)
\end{equation}
where we set $n=T$ in Algorithm~\ref{alg:1}. Compared to the strategy that predicts the majority bit with payoff $|\sum_{j} b_j |$, DNP achieves an $O(\sqrt{T \log T})$ regret by setting $Z=o(1/T)$, as well as a subconstant $o(1)$ loss. Furthermore, DNP can combine the decisions of two experts by defining $b_t$ as the difference between the losses of experts and restricting $c_t \in [0,1]$. For the general case of $N$ experts, we can use multiple DNP to aggregate experts' decisions one by one. 

\citet[Lemma 18]{DNP:2010:Arxiv} also investigate the discounted payoff $\sum_{t=1}^T \rho^{T-t} c_t b_t$, and propose a conservative updating rule to control the value of the deviation $x_t$. To be specific, Line~5 of Algorithm~\ref{alg:1} is replaced by
\begin{algorithmic}
\IF{$|x_t| < U(n)$ or $g(x_t)b_t <0$}
\STATE Set $x_{t+1}=\rho x_t + b_t$
\ELSE
\STATE Set $x_{t+1}=\rho x_t$
\ENDIF
\end{algorithmic}
where $U(n)=O(\sqrt{n \log (1/Z)})$ is a constant such that $g(x)=1$ for $|x| \geq U(n)$. It can be seen that the current bit $b_t$ is utilized to update $x_t$ only when the confidence of the algorithm is low  or  the algorithm predicts incorrectly. Then, they demonstrate that the discounted payoff is on the order of $O(-Z/(1-\rho))$  \citep[Lemma 18]{DNP:2010:Arxiv}.  Similarly, Discounted-Normal-Predictor with conservative updating can be applied to the problem of learning with expert advice. 

Following their work, \citet{pmlr-v98-daniely19a} refine the analysis of DNP to support the switching cost and the adaptive regret. They slightly modify the confidence function as 
\begin{equation} \label{eqn:confidence:new}
g(x)=\Pi_{[0,1]} \left[\tg(x) \right],\quad \tg(x)=\sqrt{\frac{n}{8}} Z \cdot \erf\left( \frac{x}{\sqrt{8 n}} \right) e^{\frac{x^2}{16 n}}
\end{equation}
where $\Pi_{[0,1]} [ \cdot ]$ denotes the projection operation onto the set $[0,1]$, and the error function is redefined as $\erf(x) = \int_0^x e^{-s^2/2} d s$. \citet[Theorem~10]{pmlr-v98-daniely19a} demonstrates that their refined DNP with projection operation attains similar regret bounds to that of \eqref{eqn:DNP:main:thm} even in the presence of switching costs. Moreover, they also extend their algorithm to support adaptive regret. Subsequently, \citet{Smooth:DNP:NeurIPS} analyze the performance of DNP with conservative updating (DNP-cu) in the context of OCO, and propose Smoothed OGD (SOGD) algorithm, where multiple instances of OGD with different step sizes are created and aggregated sequentially using DNP-cu. Their analysis shows that SOGD achieves nearly-optimal bounds for adaptive regret and dynamic regret, in the presence of switching cost. However, both of their methods do not consider discounted regret.

\begin{algorithm}[t]
\caption{Discounted-Normal-Predictor with conservative updating (DNP-cu)}
\begin{algorithmic}[1]
\REQUIRE Two parameters: $\rho$ and $Z$
\STATE Set $x_1=0$, $n=1/(1-\rho)$, and $U(n)=\tg^{-1}(1)$
\FOR{$t=1,\ldots,T$}
\STATE Predict $g(x_t)$ where $g(\cdot)$ is defined in (\ref{eqn:confidence:new})
\STATE Receive $b_t$
\IF{$x_t \in [0, U(n)]$ or $x_t <0 \&  b_t > 0$ or $x_t >U(n) \& b_t < 0$}
\STATE Set $x_{t+1}=\rho x_t + b_t$
\ELSE
\STATE Set $x_{t+1}=\rho x_t$
\ENDIF
\ENDFOR
\end{algorithmic}
\label{alg:2}
\end{algorithm}

\section{Main Results} \label{sec:main}
In this section, we present the theoretical results of this paper, including OGD for discounted OCO and uniform discounted regret across a continuous interval. 

\subsection{Online Gradient Descent for Discounted OCO}
We investigate online gradient descent (OGD) with constant step size \citep{zinkevich-2003-online} for discounted OCO.  OGD performs gradient descent to update the current solution $\w_t$:
\begin{equation} \label{eqn:ogd}
\w_{t+1} = \Pi_{\W}\big[\w_t - \eta \nabla f_t(\w_t)\big]
\end{equation}
where $\eta>0$ is the step size, and $\Pi_{\W}[\cdot]$ denotes the projection onto $\W$. In the following, we present the theoretical guarantee of OGD for discounted OCO. 

\begin{thm} \label{thm:OGD} Under Assumptions~\ref{ass:1}, \ref{ass:2} and \ref{ass:3}, for any $ \w \in \W$,  OGD  satisfies
\[
\sum_{t=1}^T  \lambda^{T-t}  f_t(\w_t)   -   \sum_{t=1}^T  \lambda^{T-t}  f_t(\w)  \leq  \frac{DG \sqrt{2}}{\sqrt{1-\lambda}}
\]
where we set $\eta = D\sqrt{2(1-\lambda)}/G$. 
\end{thm}
\paragraph{Remark:} Theorem~\ref{thm:OGD} shows that OGD achieves an $O(1/\sqrt{1 - \lambda})$ bound for $\lambda$-discounted regret, which is the same order as the $O(1/\sqrt{1 - \lambda^2})$ result established by \citet[Theorem 6]{pmlr-v235-zhang24e} because $1-\lambda<1-\lambda^2<2(1-\lambda)$. It is worth mentioning that both the step size and the upper bound are independent of the total iterations $T$, thus $O(1/\sqrt{1 - \lambda})$ holds uniformly over time. 

\subsection{Uniform Discounted Regret Across a Continuous Interval}
In this subsection, we focus on the more challenging case with an unknown discount factor. We begin by discussing the range of $\lambda$ values that are of interest. With a discount factor $0<\lambda<1$, the effective window size is essentially $\frac{1}{1-\lambda}$. For a $T$-round game, it is natural to require
\[
\frac{1}{1-\lambda} \in \left[\tau , T \right]
\]
where $\tau$ is a minimal window length introduced for technical reasons,  implying~\footnote{While there is no particular reason to avoid setting the upper bound of $\lambda$ to $1/T^\alpha$ for some $\alpha \geq 1$, we focus on $1/T$ for simplicity.}
\begin{equation} \label{eqn:lbd:range}
\lambda \in \left[1-\frac{1}{\tau}, 1-\frac{1}{T} \right].
\end{equation}

As shown in Theorem~\ref{thm:OGD}, OGD with a suitable step size can minimize the discounted regret for a particular value of $\lambda$. In order to handle all possible values of $\lambda$ in (\ref{eqn:lbd:range}), we discretize the interval $[1 - 1/\tau, 1 - 1/T]$ by introducing the following set:
\begin{equation}\label{eqn:eta:set}
\MS=\left\{ 1-\frac{1}{T}, 1-\frac{2}{T}, \ldots, 1-\frac{2^N}{T}\right\}, \textrm{ where } N= \left\lceil \log_2 \frac{T}{\tau} \right\rceil
\end{equation} 
which covers the range of discount factor values. Then, for each discount factor $\lambda_i = 1 - 2^{i - 1}/T \in \MS$, we create an instance of OGD, denoted by $\A_i$. According to Theorem~\ref{thm:OGD}, the step size of $\A_i$ is set as
\begin{equation}\label{eqn:step:etai}
\eta_i=  \frac{D\sqrt{2(1-\lambda_i)}}{G}=\frac{D}{G} \sqrt{\frac{2^i}{T}}
\end{equation} 
so that it achieves the optimal $\lambda_i$-discounted regret. As discussed in Section~\ref{sec:tc}, most adaptive methods are unable to combine these experts' decisions to attain optimal $\lambda$-discounted regret with an unknown discount factor. To address this issue, inspired by~\citet{Smooth:DNP:NeurIPS}, we choose Discounted-Normal-Predictor with conservative updating (DNP-cu)~\citep{DNP:2010:Arxiv} (summarized in Algorithm~\ref{alg:2}) as the meta-algorithm to sequentially aggregate the decisions from these multiple experts.

Before describing the specific algorithm, we first present the performance of DNP-cu (i.e., Algorithm~\ref{alg:2}) under the discounted payoff setting. Although \citet[Theorem 1]{Smooth:DNP:NeurIPS} have studied the properties of DNP-cu, their analysis is restricted to the standard payoff. 
\begin{thm} \label{thm:1} 
Suppose $Z \leq \frac{1}{e}$, $n \geq \max\{8e, 16\log \frac{1}{Z}\}$ and $U(n)\geq 22$. For any bit sequence $b_1,\ldots,b_T$ where $|b_t| \leq 1$, and any $\eta \geq \rho=1-1/n$,  the discounted payoff of Algorithm~\ref{alg:2} satisfies
\begin{equation} \label{eqn:alg2:lower3}
\sum_{t=1}^{T}  \eta^{T-t}  g(x_t) b_t  \geq  -\frac{Z}{2(1-\eta)}.
\end{equation}
Furthermore, it also satisfies
\begin{equation} \label{eqn:alg2:lower4}
\sum_{t=1}^{T}  \rho^{T-t}  g(x_t) b_t  \geq \sum_{t=1}^{T}  \rho^{T-t}   b_t  -\frac{Z}{2(1-\rho)} - U(n) - 1
\end{equation}
where  
\begin{equation} \label{eqn:u:tau}
U(n)=\tg^{-1}(1) \leq \sqrt{16n \log \frac{1}{Z}}.
\end{equation}
\end{thm}
\paragraph{Remark:} Theorem~\ref{thm:1} indicates that DNP-cu can effectively control the discounted regret. First, \eqref{eqn:alg2:lower4} shows that DNP-cu is able to support the  $\lambda$-discounted regret when $\rho=\lambda$. Furthermore, \eqref{eqn:alg2:lower3} reveals that while DNP-cu operates with a discount factor $\rho$, it can also provide a discounted payoff guarantee for a different discount factor $\eta$, provided that $\eta \geq \rho$. Therefore, we can exploit \eqref{eqn:alg2:lower3} and \eqref{eqn:alg2:lower4}  to enable aggregation of two experts operating under the discounted regret with different discount factors, as specified below.

\begin{algorithm}[t]
\caption{Combiner}
\begin{algorithmic}[1]
\REQUIRE Two parameters: $\rho$ and $Z$
\REQUIRE Two algorithms: $\A_1$ and $\A_2$
\STATE Let $\D$ be an instance of DNP-cu, i.e., Algorithm~\ref{alg:2}, with parameter $\rho$ and $Z$
\STATE Receive $\w_{1,1}$ and $\w_{1,2}$ from $\A_1$ and $\A_2$ respectively
\STATE Receive the prediction $\omega_{1}$ from $\D$
\FOR{$t=1,\ldots,T$}
\STATE Predict $\w_t$ according to \eqref{eqn:weight:combin}
\STATE Send the loss function $f_t(\cdot)$ to $\A_1$ and $\A_2$
\STATE Receive $\w_{t+1,1}$ and $\w_{t+1,2}$ from $\A_1$ and $\A_2$ respectively
\STATE Send the real bit $\ell_t$ in \eqref{eqn:ell:definition} to $\D$
\STATE Receive the prediction $\omega_{t+1}$ from $\D$
\ENDFOR
\end{algorithmic}
\label{alg:3}
\end{algorithm}

Algorithm~\ref{alg:3}, referred to as Combiner, serves as a meta-algorithm to aggregate the outputs of two OGD experts. Let $\A_1$ and $\A_2$ denote two OGD algorithm, and let $\w_{t,1}$ and $\w_{t,2}$ be their respective predictions at round $t$. Combiner generates a convex combination of $\w_{t,1}$ and $\w_{t,2}$ as its output:
\begin{equation} \label{eqn:weight:combin}
\w_t= (1-\omega_t) \w_{t,1} +  \omega_t  \w_{t,2}
\end{equation}
where the weight $\omega_t\in[0,1]$. By the convexity of $f_t(\cdot)$, we have
\[
f_t(\w_t) \leq  (1-\omega_t)f_t(\w_{t,1}) +  \omega_t f_t(  \w_{t,2}).
\]
Then, it is straightforward to verify that the $\lambda_1,\lambda_2$-discounted regret of Combiner with respect to $\A_1$ and $\A_2$ can be bounded as follows:
\begin{align}
\sum_{t=1}^T  \lambda^{T-t}_1  f_t(\w_t) - \sum_{t=1}^T  \lambda^{T-t}_1  f_t(\w_{t,1}) \leq&   -GD\sum_{t=1}^T  \lambda^{T-t}_1 \omega_t \ell_t,  \label{eqn:com:reg:1}\\
\sum_{t=1}^T  \lambda^{T-t}_2  f_t(\w_t) - \sum_{t=1}^T  \lambda^{T-t}_2  f_t(\w_{t,2}) \leq & - GD\sum_{t=1}^T  \lambda^{T-t}_2 \left(\omega_t \ell_t -\ell_t \right) \label{eqn:com:reg:2}
\end{align}
where we define
\begin{equation}  \label{eqn:ell:definition}
\ell_t =\frac{f_t(\w_{t,1})  -  f_t(\w_{t,2})}{GD} \overset{(\ref{eqn:function:range})}{\in} [-1, 1].
\end{equation}
To determine the weight, we pass $\ell_t$ to DNP-cu and set $\omega_t$ as its output. By the theoretical guarantee of DNP-cu in Theorem~\ref{thm:1}, we can use \eqref{eqn:alg2:lower3} and \eqref{eqn:alg2:lower4} to upper bound  the discounted regret \eqref{eqn:com:reg:1} and \eqref{eqn:com:reg:2}, respectively. Notably, since the discounted payoff \eqref{eqn:alg2:lower3} in Theorem~\ref{thm:1} supports arbitrary discount factor $\eta\geq \rho$, we are able to successfully aggregate the decisions of two experts for different discounted regret measurements. Specifically, let $\A_1$ and $\A_2$ be OGD algorithms for discount factors $\lambda_1$ and $\lambda_2$ respectively, with $\w_{t,1}$ and $\w_{t,2}$ as their decisions. To combine their decisions, we employ DNP-cu with $\rho=\lambda_2$ to obtain:
\begin{align*}
\sum_{t=1}^T  \lambda^{T-t}_1  f_t(\w_t) - \sum_{t=1}^T  \lambda^{T-t}_1  f_t(\w_{t,1}) \overset{\eqref{eqn:alg2:lower3},\eqref{eqn:com:reg:1}}{\leq}&    \frac{GDZ}{2(1-\lambda_1)} ,  \\
\sum_{t=1}^T  \lambda^{T-t}_2  f_t(\w_t) - \sum_{t=1}^T  \lambda^{T-t}_2  f_t(\w_{t,2})  \overset{\eqref{eqn:alg2:lower4},\eqref{eqn:com:reg:2}}{\leq} & GD \left( \frac{Z}{2(1-\lambda_2)} + U(n)+1 \right)
\end{align*}
where $\w_t$ is the combined output of $\w_{t,1}$ and $\w_{t,2}$, and we require $\eta=\lambda_1\geq \lambda_2$ by Theorem~\ref{thm:1}.

\begin{algorithm}[t]
\caption{Smoothed OGD (SOGD)}
\begin{algorithmic}[1]
\REQUIRE Two parameters: $N$ and $Z$
\STATE Set $\B_0$ be any baseline
\FOR{$i=1,\ldots,N+1$}
\STATE Let $\A_i$ be an instance of OGD with step size $\eta_i=\frac{D}{G} \sqrt{\frac{2^i}{T}}$ 
\STATE Let $\B_i$ be an instance of Combiner, i.e., Algorithm~\ref{alg:3} which combines $\B_{i-1}$ and $\A_i$ with parameters $\lambda_i=1 - 2^{i - 1}/T$ and $Z$
\ENDFOR
\FOR{$t=1,\ldots,T$}
\STATE Run $\B_1, \ldots, \B_{N+1}$ sequentially for one step each
\STATE Output the solution of $\B_{N+1}$, denoted by $\w_t$
\ENDFOR
\end{algorithmic}
\label{alg:4}
\end{algorithm}

\paragraph{Remark:} By combining the above inequalities with optimal $\lambda_i$-discounted regret achieved by the $i$-th expert, we demonstrate that when the discount factor $\lambda\in \{\lambda_1,\lambda_2\}$, our method can achieve the optimal $O(\sqrt{1/(1-\lambda)})$ bound for $\lambda$-discounted regret without requiring knowledge of the exact value of the discount factor. 

To establish uniform discounted regret over the range of discount factors specified in \eqref{eqn:lbd:range}, we construct multiple experts by running OGD with different step sizes in \eqref{eqn:eta:set}, and apply Combiner to sequentially aggregate these algorithms. For each $i \in [N+1]$, we create an instance of Combiner, denoted by $\B_i$, to combine $\B_{i-1}$ and $\A_i$, where the discount factor of $\B_i$ is set to $\lambda_i$, matching that of $\A_i$. Initially, we set $\B_0$ as any baseline, such as an algorithm that predicts a fixed point in $\W$. At each round $t$, we sequentially run $\B_1, \ldots, \B_{N+1}$ for one step each, and output the solution of $\B_{N+1}$. It is important to construct the sequence of algorithms $\A_i/\B_i$ in \emph{descending} order of their associated discount factors, as this ordering is crucial for analyzing the overall discounted regret of the algorithm. Finally, we leverage a technical lemma (\citep[Lemma~20]{DNP:2010:Arxiv}) to extend the results from a discrete set to a continuous interval. Following \citet{Smooth:DNP:NeurIPS}, we refer to this algorithm as Smoothed OGD (SOGD), and summarize it in Algorithm~\ref{alg:4}.

We have the following theoretical guarantee regarding the discounted regret of SOGD.
\begin{thm} \label{thm:main} Suppose $\tau \geq \max\{16e, 32\log \frac{1}{Z}\}$ and set $Z=1/T$. Under Assumptions~\ref{ass:1}, \ref{ass:2} and \ref{ass:3},  for any $\lambda \in [1-1/\tau, 1-1/T]$, Algorithm~\ref{alg:4} satisfies
\[
 \begin{split}
\sum_{t=1}^T  \lambda^{T-t}  f_t(\w_t)   -   \sum_{t=1}^T  \lambda^{T-t}  f_t(\w)  \leq \frac{2GD}{\sqrt{1-\lambda}} \left( 4 \sqrt{ \log \frac{1}{Z}} + \sqrt{2}\right)  + \frac{GD(N+1)Z}{1-\lambda}+2GD
\end{split}
\]
for any $\w \in \W$, where $N= \left\lceil \log_2 \frac{T}{\tau} \right\rceil$.
\end{thm} 

\paragraph{Remark:} Theorem~\ref{thm:main} shows that SOGD achieves an $O(\sqrt{\log T / (1-\lambda)})$ bound for $\lambda$-discounted regret, holding simultaneously for all $\lambda \in [1-1/\tau, 1-1/T]$. Compared to the $O\left(1/\sqrt{1-\lambda}\right)$ guarantee in Theorem~\ref{thm:OGD} for a known discount factor $\lambda$, this bound incurs an additional $O(\sqrt{\log T})$ factor, reflecting the cost of adaptivity to the discount factor.

\section{Analysis}\label{sec:ana}
In this section, we present the analysis of all theorems.

\subsection{Proof of Theorem~\ref{thm:OGD}}
The proof is similar to that of  \citet[Theorem 6]{pmlr-v235-zhang24e}, except that we adopt a different step size. From Jensen's inequality and the standard analysis of OGD, we have
\[
\begin{split}
 &\sum_{t=1}^T  \lambda^{T-t}  f_t(\w_t)   -   \sum_{t=1}^T  \lambda^{T-t}  f_t(\w) \\
\leq & \sum_{t=1}^T  \lambda^{T-t}  \langle \nabla f_t(\w_t), \w_t -\w \rangle \\
\leq &\sum_{t=1}^T   \frac{\lambda^{T-t} }{2 \eta} \left( \|\w_t -\w \|^2 - \|\w_{t+1}-\w\|^2 \right) +  \sum_{t=1}^T \frac{\eta \lambda^{T-t}}{2} \| f_t(\w_t)\|^2 \\
\leq &  \frac{\lambda^{T-1} }{2 \eta}  \|\w_1 -\w \|^2 +  \sum_{t=2}^{T} \left(\frac{\lambda^{T-t} }{2 \eta} -\frac{\lambda^{T-(t-1)} }{2 \eta}\right) \|\w_t -\w \|^2 +   \frac{\eta }{2} \sum_{t=1}^T  \lambda^{T-t} \| f_t(\w_t)\|^2  \\
\leq & \frac{\lambda^{T-1} D^2}{2 \eta} + \frac{(1-\lambda) D^2}{2 \eta} \sum_{t=2}^{T} \lambda^{T-t} +   \frac{\eta G^2}{2} \sum_{t=1}^T  \lambda^{T-t} \\
\leq & \frac{\lambda^{T-1} D^2}{2 \eta} + \frac{(1-\lambda) D^2}{2 \eta} \frac{1}{1-\lambda} +   \frac{\eta G^2}{2} \frac{1}{1-\lambda} \leq \frac{D^2}{ \eta} +   \frac{\eta G^2}{2(1-\lambda)}= \frac{DG \sqrt{2}}{\sqrt{1-\lambda}} 
\end{split}
\]
where in the last step we set 
\[
\eta = \frac{D\sqrt{2(1-\lambda)}}{G}.
\]

\subsection{Proof of Theorem~\ref{thm:1}}
The proof builds on the approach of \citet[Theorem 1]{Smooth:DNP:NeurIPS}, and we highlight the main difference below.

Following the arguments of \citet{DNP:2010:Arxiv} and \citet{Smooth:DNP:NeurIPS}, we analyze the payoff of Algorithm~\ref{alg:2} by leveraging Algorithm~\ref{alg:1}. This is because the update rule of Algorithm~\ref{alg:1} is simpler, making the analysis more tractable. Specifically, we construct the following bit sequence
\[
\tb_t = \left\{
            \begin{array}{ll}
              b_t, & \hbox{if Line 6 of Algorithm~\ref{alg:2} is executed at round $t$}; \\
              0, & \hbox{otherwise.}
            \end{array}
          \right.
\]
It is straightforward to verify that the prediction $g(x_t)$, as well as the deviation $x_t$, generated by Algorithm~\ref{alg:2} on the bit sequence $b_1, \ldots, b_T$, is exactly the same as that of Algorithm~\ref{alg:1} on the transformed sequence $\tb_1, \ldots, \tb_T$. In the following, we first establish the theoretical guarantee for Algorithm~\ref{alg:1} on the transformed sequence and then translate this result to the payoff of Algorithm~\ref{alg:2} on the original sequence. We establish the following theorem for the discounted payoff of Algorithm~\ref{alg:1}.
\begin{thm} \label{thm:2} Suppose $Z \leq \frac{1}{e}$ and $n \geq \max\{8e, 16\log \frac{1}{Z}\}$. For any bit sequence $b_1,\ldots,b_T$ such that $|b_t| \leq 1$, and any $\eta \geq \rho=1-1/n$,  the discounted payoff of Algorithm~\ref{alg:1} satisfies
\begin{equation} \label{eqn:alg1:lower1}
\sum_{t=1}^{T}  \eta^{T-t}  g(x_t) b_t  \geq  \sum_{t=1}^{T} \frac{ \eta^{T-t}}{n} \left(\frac{x_tg(x_t)}{2} - \Phi_t \right)-\frac{Z}{2(1-\eta)}
\end{equation}
where $\Phi_t = \int_0^{x_t} g(s) ds$ is the potential function. Furthermore,
\begin{equation} \label{eqn:alg1:lower2}
\sum_{t=1}^{T}  \rho^{T-t}  g(x_t) b_t  \geq \sum_{t=1}^{T}  \rho^{T-t}   b_t + \sum_{t=1}^{T} \frac{\rho^{T-t}}{n} \left(\frac{x_tg(x_t)}{2} - \Phi_t \right)-\frac{Z}{2(1-\rho)} - x_{T+1}.
\end{equation}
\end{thm}

From Theorem~\ref{thm:2}, we have
\begin{align}
\sum_{t=1}^{T}  \eta^{T-t}  g(x_t) \tb_t & \geq  \sum_{t=1}^{T} \frac{ \eta^{T-t}}{n} \left(\frac{x_tg(x_t)}{2} - \Phi_t \right)-\frac{Z}{2(1-\eta)}, \label{eqn:alg1:lower1:trans} \\
\sum_{t=1}^{T}  \rho^{T-t}  g(x_t) \tb_t  &\geq \sum_{t=1}^{T}  \rho^{T-t}   \tb_t + \sum_{t=1}^{T} \frac{\rho^{T-t}}{n} \left(\frac{x_tg(x_t)}{2} - \Phi_t \right)-\frac{Z}{2(1-\rho)} - x_{T+1}.  \label{eqn:alg1:lower2:trans}
\end{align}
From (38) of \citet{Smooth:DNP:NeurIPS}, we have
\begin{align}
g(x_k) (b_k -  \tb_k) &\geq  0,  \textrm{ if } b_k \neq \tb_k, \label{eqn:diff:bound1} \\
g(x_k) (b_k -  \tb_k) &\geq  b_k-\tb_k, \textrm{ if } b_k \neq \tb_k.  \label{eqn:diff:bound2}
\end{align}
which implies
\begin{align}
\begin{split}
\sum_{t=1}^{T}  \eta^{T-t}  g(x_t) b_t = &\sum_{t=1}^{T}  \eta^{T-t}  g(x_t) \tb_t + \sum_{t=1}^{T}  \eta^{T-t}  g(x_t) (b_t-\tb_t) \\
\overset{(\ref{eqn:alg1:lower1:trans}),(\ref{eqn:diff:bound1})}{\geq}   &\sum_{t=1}^{T} \frac{ \eta^{T-t}}{n} \left(\frac{x_tg(x_t)}{2} - \Phi_t \right)-\frac{Z}{2(1-\eta)},
\end{split} \label{eqn:alg2:lower1}\\
\begin{split}
\sum_{t=1}^{T}  \rho^{T-t}  g(x_t) b_t = &\sum_{t=1}^{T}  \rho^{T-t}  g(x_t) \tb_t + \sum_{t=1}^{T}  \rho^{T-t}  g(x_t) (b_t-\tb_t) \\
\overset{(\ref{eqn:alg1:lower2:trans}),(\ref{eqn:diff:bound2})}{\geq}   &\sum_{t=1}^{T}  \rho^{T-t}   b_t + \sum_{t=1}^{T} \frac{\rho^{T-t}}{n} \left(\frac{x_tg(x_t)}{2} - \Phi_t \right)-\frac{Z}{2(1-\rho)} - x_{T+1}.\label{eqn:alg2:lower2}
\end{split}
\end{align}

To simplify the term involving $\frac{x_tg(x_t)}{2} - \Phi_t$, we make use of the following property of Algorithm~\ref{alg:1}   \citep[(33)]{Smooth:DNP:NeurIPS}
\begin{equation}\label{eqn:xt:range}
-1 \leq x_t  \leq U(n) + 1, \ \forall t \geq 1
\end{equation}
and establish the following lemma.
\begin{lem} \label{lem:poten} Suppose $Z \leq \frac{1}{e}$, $n \geq \max\{8e, 16\log \frac{1}{Z}\}$ and $U(n)\geq 20$. Under the condition in (\ref{eqn:xt:range}), we have
\begin{equation} \label{eqn:pot:upper}
\Phi_t = \int_0^{x_t} g(s) ds \leq \frac{x_tg(x_t)}{2} .
\end{equation}
\end{lem}
Finally, we obtain (\ref{eqn:alg2:lower3}) by combining (\ref{eqn:alg2:lower1}) and (\ref{eqn:pot:upper}), and obtain (\ref{eqn:alg2:lower4}) by combining (\ref{eqn:alg2:lower2}), (\ref{eqn:xt:range}) and (\ref{eqn:pot:upper}).

\subsection{Proof of Theorem~\ref{thm:main}}
We start by presenting the discounted regret of each OGD algorithm, namely $\A_i$. Let $\w_{t,i}$ be the output of $\A_i$ at round $t$, and  recall that the step size is set according to (\ref{eqn:step:etai}). From Theorem~\ref{thm:OGD}, we have
\begin{equation}\label{eqn:ai:regret}
\sum_{t=1}^T  \lambda_i^{T-t}  f_t(\w_{t,i})   -   \sum_{t=1}^T  \lambda_i^{T-t}  f_t(\w)  \leq  \frac{DG \sqrt{2}}{\sqrt{1-\lambda_i}}
\end{equation}
for all $\w \in \W$, where $\lambda_i = 1 - 2^{i - 1}/T $.

Next, we proceed to bound the discounted regret of SOGD at each discount factor $\lambda_i$, where $i \in [N+1]$. To this end, we make use of the theoretical guarantee of DNP-cu stated in Theorem~\ref{thm:1}. Note that the conditions of Theorem~\ref{thm:1} impose a lower bound on the window size $n$. Therefore, to apply this result, we need to ensure that the minimal window size appearing in Algorithm~\ref{alg:4} satisfies
\begin{equation} \label{eqn:condition:tau}
\min_{i\in[N+1]} \frac{1}{1-\lambda_i}= \frac{T}{2^N} \geq \max \left\{8e, 16\log \frac{1}{Z}\right\}.
\end{equation}
Since 
\[
\frac{T}{2^N} \geq \frac{T}{2^{ 1+\log_2 \frac{T}{\tau}}} = \frac{\tau}{2} ,
\]
(\ref{eqn:condition:tau}) holds when $ \tau \geq \max\{16e, 32\log \frac{1}{Z}\}$. Furthermore, we also require $U(n)\geq 22$ in Theorem~\ref{thm:1}. We now proceed to estimate a lower bound for $U(n)$. Since $g(x)$ is a monotonically increasing function on $[0,U(n)]$, it follows that if we can find a point $x^\prime$ such that $g(x^\prime)<1$, then we must have $x^\prime < U(n)$. Setting $x=4\sqrt{n}$, we have
\begin{equation*}
    g(4\sqrt{n}) = \sqrt{\frac{n}{8}}Z\cdot \erf \left( \frac{1}{\sqrt{2}} \right)e \leq \frac{\sqrt{\pi}}{4\sqrt{2}} e \approx 0.8517
\end{equation*}
where the last step is because we set $Z=1/T$ and $n=\tau\leq T$, and property of the error function. Since $g(4\sqrt{n}) <1$, we have 
\begin{equation*}
    U(n)\geq 4\sqrt{n} \geq 4\sqrt{32}\geq 22. 
\end{equation*}

Denote the output of $\B_i$ at round $t$ by $\v_{t,i}$, and observe that the final output is $\w_t = \v_{t,N+1}$. Then, we decompose the $\lambda_i$-discounted regret of SOGD as
\begin{equation} \label{eqn:regret:deco}
\begin{split}
&\sum_{t=1}^T  \lambda_i^{T-t}  f_t(\w_{t})   -   \sum_{t=1}^T  \lambda_i^{T-t}  f_t(\w) = \sum_{t=1}^T  \lambda_i^{T-t}  f_t(\v_{t,N+1})   -   \sum_{t=1}^T  \lambda_i^{T-t}  f_t(\w) \\
=& \sum_{k=i+1}^{N+1}  \underbrace{\sum_{t=1}^T  \left(\lambda_i^{T-t}  f_t(\v_{t,k}) - \lambda_i^{T-t}  f_t(\v_{t,k-1})\right)}_{\alpha_k}  +\underbrace{\sum_{t=1}^T  \lambda_i^{T-t}  f_t(\v_{t,i})- \sum_{t=1}^T  \lambda_i^{T-t}  f_t(\w_{t,i})}_{\beta}  \\
&+\underbrace{\sum_{t=1}^T  \lambda_i^{T-t}  f_t(\w_{t,i})   -   \sum_{t=1}^T  \lambda_i^{T-t}  f_t(\w) }_{\gamma}
\end{split}
\end{equation}
where $\alpha_k$ denotes the $\lambda_i$-discounted regret of $\B_k$ with respect to $\B_{k-1}$, $\beta$ denotes the $\lambda_i$-discounted regret of $\B_i$ with respect to $\A_i$, and $\gamma$ denotes the $\lambda_i$-discounted regret of $\A_i$. We can directly bound $\gamma$ using (\ref{eqn:ai:regret}), so we next focus on bounding $\alpha_k$ for $k = i+1, \ldots, N+1$, and $\beta$.

Recall that $\B_k$ invokes DNP-cu with discount factor $\lambda_k$ to aggregate $\B_{k-1}$ and $\A_k$. Define
\[
\ell_{t,k} =  \frac{f_t(\v_{t,k-1})  -  f_t(\w_{t,k})}{GD} . 
\]
Following the derivation of (\ref{eqn:com:reg:1}), we have 
\begin{equation} \label{eqn:alpha_k1}
\alpha_k=\sum_{t=1}^T  \lambda_i^{T-t}  f_t(\v_{t,k}) -\lambda_i^{T-t}  f_t(\v_{t,k-1}) \leq  -GD\sum_{t=1}^T  \lambda_i^{T-t} \omega_{t,k} \ell_{t,k} 
\end{equation}
where $\omega_{t,k}$ is  the output of the $k$-th instance of DNP-cu at round $t$. Since the discount factors are set in descending order in Algorithm~\ref{alg:4}, it holds that
\[
\lambda_i > \lambda_k, \textrm{ for all }  k=i+1,\ldots, N+1.
\]
Thus, we can use (\ref{eqn:alg2:lower3}) in Theorem~\ref{thm:1} to upper bound the RHS of (\ref{eqn:alpha_k1}), yielding 
\begin{equation} \label{eqn:alpha_k2}
\alpha_k \overset{(\ref{eqn:alg2:lower3}),(\ref{eqn:alpha_k1})}{\leq}  \frac{GDZ}{2(1-\lambda_i)} .
\end{equation}

Similarly, following the derivation of (\ref{eqn:com:reg:2}), we have
\begin{equation} \label{eqn:beta}
\begin{split}
\beta=&\sum_{t=1}^T  \lambda_i^{T-t}  f_t(\v_{t,i})- \sum_{t=1}^T  \lambda_i^{T-t}  f_t(\w_{t,i}) \leq  - GD\sum_{t=1}^T  \lambda_i^{T-t} \left(\omega_{t,i} \ell_{t,i} -\ell_{t,i} \right) \\ 
\overset{(\ref{eqn:alg2:lower4})}{\leq} & GD \left( \frac{Z}{2(1-\lambda_i)} + U\left(\frac{1}{1-\lambda_i}\right) +1\right)   \overset{(\ref{eqn:u:tau})}{\leq} GD \left( \frac{Z}{2(1-\lambda_i)} +\sqrt{16 \frac{1}{1-\lambda_i} \log \frac{1}{Z}} +1\right).
\end{split}
\end{equation}
Substituting  (\ref{eqn:ai:regret}), (\ref{eqn:alpha_k2}) and (\ref{eqn:beta}) into (\ref{eqn:regret:deco}), we obtain
\begin{equation}\label{eqn:regret:dis}
\sum_{t=1}^T  \lambda_i^{T-t}  f_t(\w_{t})   -   \sum_{t=1}^T  \lambda_i^{T-t}  f_t(\w) \leq    \frac{GD}{\sqrt{1-\lambda_i}} \left( 4 \sqrt{ \log \frac{1}{Z}} + \sqrt{2}\right)  + \frac{GD(N+1)Z}{2(1-\lambda_i)}+GD
\end{equation}
for all  $i=1,\ldots, N+1$. 

So far, we have established discounted regret guarantees for each discount factor $\lambda_i$, where $i \in [N+1]$. Next, we extend these results to the continuous range of $\lambda$ values specified in (\ref{eqn:lbd:range}). To this end, we make use of the following lemma \citep[Lemma 20]{DNP:2010:Arxiv}.
\begin{lem}\label{lem:extend} Given a sequence $s_1,\ldots,s_T$, and a discount factor $\lambda<1$, define the $\lambda$-smoothed average at time $T$ by
\[
s_T^\lambda=(1-\lambda) \sum_{t=1}^T \lambda^{T-t} s_t. 
\]
Then, for any $\lambda_1 > \lambda_2$, the $\lambda_1$-smoothed average at time $T$ is a convex combination of $\lambda_2$-smoothed average at time $j\leq T$:
\[
s_T^{\lambda_1} = \frac{1-\lambda_1}{1-\lambda_2}s_T^{\lambda_2} + \sum_{j <T} \frac{(1-\lambda_1) (\lambda_1-\lambda_2) \lambda_1^{T-j-1}}{1-\lambda_2} s_{T-j}^{\lambda_2}
\]
where the coefficients on the RHS are nonnegative and sum to $1$. 
\end{lem}

According to our construction in (\ref{eqn:eta:set}), for each $\lambda \in  [1-1/\tau, 1-1/T]$, there exists an index $i \in [N]$ such that
\[
\lambda_{i+1} = 1-\frac{2^{i}}{T} \leq \lambda \leq \lambda_i =1-\frac{2^{i-1}}{T} 
\]
implying
\begin{equation}\label{eqn:ine:lamda}
\frac{1}{1-\lambda_{i+1}} \leq \frac{1}{1-\lambda}, \textrm{ and } \frac{1-\lambda_{i+1}}{1-\lambda} \leq 2.
\end{equation}
From Lemma~\ref{lem:extend}, we know that
\[
(1-\lambda)  \sum_{t=1}^T  \lambda^{T-t} \big[ f_t(\w_{t})   -    f_t(\w) \big]
\]
can be expressed as a convex combination of 
\[
(1-\lambda_{i+1}) \sum_{t=1}^T  \lambda_{i+1}^{T-t} \big[ f_t(\w_{t})   -    f_t(\w) \big],  (1-\lambda_{i+1}) \sum_{t=1}^{T-1}  \lambda_{i+1}^{T-1-t} \big[ f_t(\w_{t})   -    f_t(\w) \big], \ldots .
\]
As a result
\[
\begin{split}
&(1-\lambda)  \sum_{t=1}^T  \lambda^{T-t} \big[ f_t(\w_{t})   -    f_t(\w) \big] \\
\leq & (1-\lambda_{i+1}) \max_{j \in[T]}\sum_{t=1}^j  \lambda_{i+1}^{j-t} \big[ f_t(\w_{t})   -    f_t(\w) \big] \\
\overset{(\ref{eqn:regret:dis})}{\leq} & (1-\lambda_{i+1}) \left[ \frac{GD}{\sqrt{1-\lambda_{i+1}}} \left( 4 \sqrt{ \log \frac{1}{Z}} + \sqrt{2}\right)  + \frac{GD(N+1)Z}{2(1-\lambda_{i+1})}+GD \right]
\end{split}
\]
where we use the fact that (\ref{eqn:regret:dis}) holds for any integer $T$. Thus, we have
\[
\begin{split}
&\sum_{t=1}^T  \lambda^{T-t} \big[ f_t(\w_{t})   -    f_t(\w) \big] \\
\leq {} & \frac{1-\lambda_{i+1}}{1-\lambda} \left[ \frac{GD}{\sqrt{1-\lambda_{i+1}}} \left( 4 \sqrt{ \log \frac{1}{Z}} + \sqrt{2}\right)  + \frac{GD(N+1)Z}{2(1-\lambda_{i+1})}+GD \right] \\
\overset{(\ref{eqn:ine:lamda})}{\leq} {} & \frac{2GD}{\sqrt{1-\lambda}} \left( 4 \sqrt{ \log \frac{1}{Z}} + \sqrt{2}\right)  + \frac{GD(N+1)Z}{1-\lambda}+2GD.
\end{split}
\]
Finally, we set $Z=1/T$ to finish the proof. 

\subsection{Proof of Theorem~\ref{thm:2}}
The overall proof strategy is similar to that of  \citet[Theorem 5]{Smooth:DNP:NeurIPS}, with the key distinction being that we consider the discounted payoff rather than the standard payoff. Moreover, our proof is more concise, as we focus on the interval $[1, T]$ and omit the switching cost. Building upon their analysis while adapting it to our setting, we directly leverage several intermediate results to avoid redundancy.

From (70) of \citet{Smooth:DNP:NeurIPS}, we know the difference between any two consecutive deviations is bounded by $2$, i.e.,
\begin{equation} \label{eqn:xt:difference}
|x_t - x_{t+1}|  \leq 2.
\end{equation}
Let $\ind(x)$ be the indicator function of the interval $[0, U(n)+2]$. We have \citep[(74)]{Smooth:DNP:NeurIPS}
\begin{equation} \label{eqn:grad:bound}
4 \max_{s \in [x_t, x_{t+1}]} |g'(s)| \leq \frac{1}{n} x_t g(x_t) \ind(x_t)+Z.
\end{equation}
We also recall the following inequality for piecewise differentiable functions $f:[a,b] \mapsto \R$ \citep{DNP:2010:Arxiv,pmlr-v98-daniely19a}:
\begin{equation} \label{eqn:integral}
\int_a^b f(x) dx \leq  f(a)(b-a) + \max|f'(z)| \frac{1}{2} (b-a)^2 .
\end{equation}

Following the idea outlined in \citet[proof of Lemma 18]{DNP:2010:Arxiv}, we have
\begin{equation} \label{eqn:diff:poten}
\begin{split}
\Phi_{t+1} -\eta \Phi_t = & \Phi_{t+1} - \Phi_t + (1-\eta) \Phi_t =  \int_{x_t}^{x_{t+1}} g(s) ds + (1-\eta) \Phi_t\\
\overset{(\ref{eqn:integral})}{\leq} & g(x_t) ( x_{t+1}- x_{t}) + \frac{1}{2} ( x_{t+1}- x_{t})^2 \max_{s\in [x_t, x_{t+1}]} |g'(s)| + (1-\eta) \Phi_t \\
\overset{(\ref{eqn:xt:difference})}{\leq} & g(x_t) \left( -\frac{1}{n} x_t + b_t \right) + 2  \max_{s\in [x_t, x_{t+1}]} |g'(s)|+ (1-\eta) \Phi_t \\
\overset{(\ref{eqn:grad:bound})}{\leq}& g(x_t) \left( -\frac{1}{n} x_t + b_t \right) + \frac{1}{2n} x_t g(x_t) \ind(x_t)+\frac{Z}{2}+ (1-\eta) \Phi_t \\
=&g(x_t) b_t + \frac{1}{2n}  x_t g(x_t) \left( \ind(x_t)   -1 \right) +(1-\eta) \Phi_t  -\frac{1}{2 n} x_t g(x_t) +\frac{Z}{2}\\
\leq & g(x_t) b_t + \frac{1}{2n}  x_t g(x_t) \left( \ind(x_t)   -1 \right) +\frac{1}{n} \left(\Phi_t - \frac{x_t g(x_t)}{2}\right) +\frac{Z}{2}
\end{split}
\end{equation}
where the last line follows from the condition $\eta \geq 1-1/n$. Then, we have
\begin{equation} \label{eqn:sum:poten}
\begin{split}
0\leq&\Phi_{T+1} =\Phi_{T+1}- \eta^T \Phi_1 = \sum_{t=1}^{T} \eta^{T-t} \left( \Phi_{t+1} -\eta \Phi_t \right) \\
\overset{(\ref{eqn:diff:poten})}{\leq}& \sum_{t=1}^{T}  \eta^{T-t}  g(x_t) b_t + \sum_{t=1}^{T} \frac{1}{2n}  \eta^{T-t} x_t g(x_t) \left( \ind(x_t)   -1 \right) + \sum_{t=1}^{T} \frac{ \eta^{T-t}}{n} \left(\Phi_t - \frac{x_t g(x_t)}{2}\right) +\frac{Z}{2} \sum_{t=1}^{T} \eta^{T-t}\\
\leq  & \sum_{t=1}^{T}  \eta^{T-t}  g(x_t) b_t + \sum_{t=1}^{T} \frac{1}{2n}  \eta^{T-t} x_t g(x_t) \left( \ind(x_t)   -1 \right) + \sum_{t=1}^{T} \frac{ \eta^{T-t}}{n} \left(\Phi_t - \frac{x_t g(x_t)}{2}\right) +\frac{Z}{2(1-\eta)} 
\end{split}
\end{equation}
where in the first equality we use the fact that $\Phi_1=0$ since $x_1=0$. 

We proceed to bound the second term in the last line of (\ref{eqn:sum:poten}). Combining the simple observation that
\[
 x_t g(x_t) \left( 1 -\ind(x_t)  \right) \geq  0,
\]
with (\ref{eqn:sum:poten}), we obtain (\ref{eqn:alg1:lower1}). Recall that
\[
 x_{T+1}=\sum_{t=1}^{T}  \rho^{T-t}   b_t  
\]
and thus
\[
\sum_{t=1}^{T}  \rho^{T-t}  g(x_t) b_t  \overset{(\ref{eqn:alg1:lower1})}{\geq}  \sum_{t=1}^{T}  \rho^{T-t}   b_t + \sum_{t=1}^{T} \frac{\rho^{T-t}}{n} \left(\frac{x_tg(x_t)}{2} - \Phi_t \right)-\frac{Z}{2(1-\rho)} - x_{T+1}.
\]

\subsection{Proof of Lemma~\ref{lem:poten}}
We note that \citet[Lemma 22]{DNP:2010:Arxiv} have established a similar result for their confidence function in (\ref{eqn:confidence:original}), but provided only an incomplete proof. In this paper, we adopt a novel and insightful approach to provide an entirely new and complete analysis. 

\begin{figure}[t]
\begin{center}
\centerline{\includegraphics[width=.4\columnwidth]{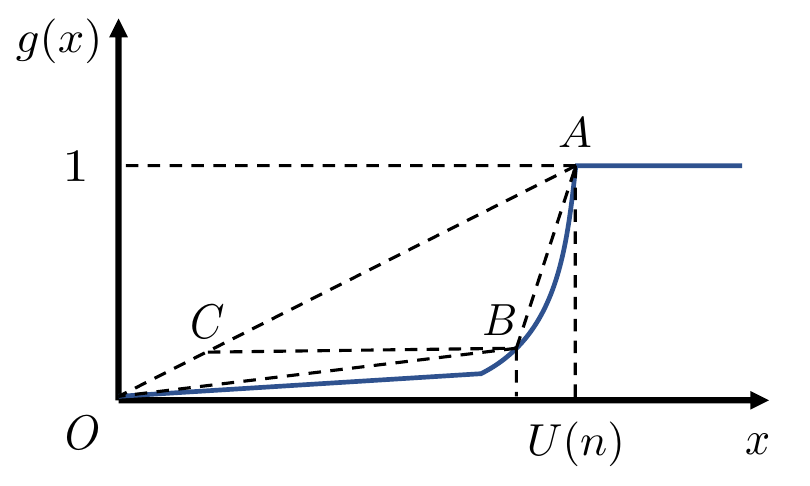}}
\caption{The confidence function $g(x)$. }
\label{fig:confidence}
\end{center}
\end{figure}

First, when $x_t \in [-1,0]$,  we have
\[
\Phi_t = 0 \leq \frac{x_tg(x_t)}{2} =0.
\]
since $g(s)=0$ for $s \in[-1,0]$. Second, when $x_t \in [0, U(n)]$, it follows from \citet[Lemma 18]{pmlr-v98-daniely19a} that $g(s)$ is convex over $[0, x_t]$. Therefore, (\ref{eqn:pot:upper}) holds by the convexity of $g(s)$. Third, when $x_t \in [U(n), U(n)+1]$, we have
\[
\Phi_t = \int_0^{x_t} g(s) ds = \int_0^{U(n)} g(s) ds + x_t-U(n)
\]
According to definition of $g(x)$, we have $\frac{x_tg(x_t)}{2}=\frac{x_t}{2}$ when $x_t \in [U(n), U(n)+1]$. Accordingly, our objective reduces to proving the following inequality:
\begin{equation*}
    \Phi_t = \int_0^{U(n)} g(s) ds + x_t-U(n) \leq \frac{x_t}{2}
\end{equation*}
for all $x_t \in [U(n), U(n)+1]$, which implies
\begin{equation}\label{eqn:goal}
    \int_0^{U(n)} g(s) ds \leq U(n) - \frac{x_t}{2}. 
\end{equation}
To establish that \eqref{eqn:goal} holds for all $x_t \in [U(n), U(n)+1]$, it suffices to prove the inequality for the maximum value of $x_t$ in this interval, that is, when $x_t=U(n)+1$. Specifically, we need to show that 
\begin{equation}\label{eqn:goal2}
    \int_0^{U(n)} g(s) ds \leq \frac{U(n)-1}{2}. 
\end{equation}
As long as \eqref{eqn:goal2} can be established, the proof of Lemma~1 is complete. However, proving \eqref{eqn:goal2} remains highly challenging. To address this, we propose a novel approach that combines analytical and geometric insights. The confidence function $g(x)$ is illustrated in Figure~\ref{fig:confidence}. 

Notably, we identify a point $B:(c,g(c))$ on the function $g(x)$. If we can show that the following inequality holds at this point, then \eqref{eqn:goal2} is immediately satisfied: 
\begin{equation*}
    \int_0^{U(n)} g(s) ds + S_{\triangle OAB} \leq \frac{U(n)}{2},\text{ and } S_{\triangle OAB}\geq \frac{1}{2}. 
\end{equation*}
From Figure~\ref{fig:confidence}, it can be seen that the area of triangle $\triangle OAB$ is equal to the sum of the areas of triangles $\triangle OBC$ and $\triangle ABC$. Therefore, we have 
\begin{equation*}
    S_{\triangle OAB} = \frac{1}{2}BC= \frac{1}{2} \left(c-U(n)\cdot g(c)\right) 
\end{equation*}
where the point  $C$ is $(U(n)\cdot g(c),g(c))$ because this point is on the line $OA:y=\frac{x}{U(n)}$. To prove \eqref{eqn:goal2}, we need to find a point $B:(c,g(c))$ that it satisfies
\begin{equation*}
    c-U(n)\cdot g(c)\geq 1.
\end{equation*}
To achieve this, we find a specific point $B$ such that $c=U(n)-8$, and we have
\begin{equation*}
    \begin{aligned}
        g(U(n)-8) &= \sqrt{\frac{n}{8}} Z\cdot \erf \left( \frac{U(n)-8}{\sqrt{8n}} \right)e^{\frac{(U(n)-8)^2}{16n}} \\
        &\leq \sqrt{\frac{n}{8}} Z\cdot \erf \left( \frac{U(n)}{\sqrt{8n}} \right)e^{\frac{(U(n)-8)^2}{16n}} \\
        & = e^{\frac{64-16U(n)}{16n}} \\
        &\leq e^{\frac{64-16U(n)}{512}}
    \end{aligned}
\end{equation*}
where the last two steps are due to the definition of $U(n)$ and $n\geq 32$. Then, we find that
\begin{equation*}
\begin{aligned}
    c-U(n)\cdot g(c) &= U(n)-8-U(n)\cdot g(U(n)-8) \\
    &\geq U(n) \left(1- e^{\frac{64-16U(n)}{512}}\right) -8 \\
    &\geq 1.4648
\end{aligned}
\end{equation*}
where the last step is due to $U(n)\geq 22$. We finish the proof. 

\section{Conclusion} \label{sec:con}
In this paper, we study discounted OCO with a discount factor $\lambda$. First, we investigate OGD with constant step size, and prove that OGD with step size $\eta=O(\sqrt{1-\lambda})$ achieves an $O(1/\sqrt{1-\lambda})$ bound for $\lambda$-discounted regret. Second, we focus on the challenging case with an unknown discount factor. To establish discounted regret for all possible factors across a continuous interval, we provide a novel analysis on the discounted payoff of DNP-cu, and show that DNP-cu enables aggregation of two experts for the discounted regret with different discount factors. Thus, we employ DNP-cu to sequentially aggregate the decisions from experts with different configurations to achieve our goal.

\bibliography{Discounted}

\end{document}